\documentclass{article}
\usepackage{ijcai22}

\usepackage{times}
\usepackage{soul}
\usepackage{url}
\usepackage[hidelinks]{hyperref}
\usepackage[utf8]{inputenc}
\usepackage[small]{caption}
\usepackage{graphicx}
\usepackage{amsmath}
\usepackage{amsthm}
\usepackage{amssymb}
\usepackage{booktabs}
\usepackage{algorithm}
\usepackage{algorithmic}
\usepackage{tikz}
\usepackage{natbib}
\usepackage{tabularx}
\usepackage{xcolor}
\usepackage{dsfont}

\newcommand*{\Scale}[2][4]{\scalebox{#1}{$#2$}}%

\newcommand{\vect}[1]{\boldsymbol{#1}}
\DeclareMathOperator*{\argmax}{arg\,max}

\title{LETS-GZSL: A Latent Embedding Model for Time Series Generalized Zero
Shot Learning}

\author{
Sathvik Bhaskarpandit$^1$
\and
Priyanka Gupta$^1$$^2$\and
Manik Gupta$^1$
\affiliations
$^1$BITS Pilani Hyderabad Campus, $^2$CVR College of Engineering\\
\emails
\{f20191200, p20190501, manik\}@hyderabad.bits-pilani.ac.in
}

\begin{document}

\maketitle

\begin{abstract}
      One of the recent developments in deep learning is generalized zero-shot learning (GZSL), which aims to recognize objects from both seen and unseen classes, when only the labeled examples from seen classes are provided. Over the past couple of years, GZSL has picked up traction and several models have been proposed to solve this problem. Whereas an extensive amount of research on GZSL has been carried out in fields such as computer vision and natural language processing, no such research has been carried out to deal with time series data. GZSL is used for applications such as detecting abnormalities from ECG and EEG data and identifying unseen classes from sensor, spectrograph and other devices' data. In this regard, we propose a Latent Embedding for Time Series - GZSL (LETS-GZSL) model that can solve the problem of GZSL for time series classification (TSC). We utilize an embedding-based approach and combine it with attribute vectors to predict the final class labels. We report our results on the widely popular UCR archive datasets. Our framework is able to achieve a harmonic mean value of at least 55\% on most of the datasets except when the number of unseen classes is greater than 3 or the amount of data is very low (less than 100 training examples). 
\end{abstract}

\section{Introduction}
Amongst the various problems related to handling time series data, time series classification (TSC) proves to be challenging due to presence of temporal ordering in the data \citep{yu2021analysis}. A data set $D = \{(x_1,y_1),(x_2,y_2),...,(x_N,y_N)\}$ is a set of pairs $(x_i,y_i)$ where $x_i=\{ x_{i,1}, x_{i,2}, ... x_{i,T} \}$ is a univariate time series, $y_i$ is its label, N is the total number of time series in the dataset and T is the total number of time stamps in a univariate time series. Given a dataset $D$, the aim of TSC is to output a label \(\hat{y}_{i}\) given a new univariate time series as input \citep{lei2020time}.

The challenge of TSC is to capture the temporal patterns that exist in the data and identify the discriminatory features to differentiate between various classes. Several distance-based methods \citep{abanda2019review}, machine learning (ML) \citep{esmael2012improving} and deep learning (DL) \citep{fawaz2019deep} models have been proposed to solve the TSC problem. Since distance-based methods had high computation costs and performed poorly with noisy time series, ML, especially DL methods gained much more focus in research. The features that are selected or extracted for building ML models greatly affect the performance,  \citep{wang2006characteristic} yet DL models provide an advantage of automated feature extraction. Several DL models have been proposed that achieve state of the art results on TSC \citep{fawaz2019deep}. Furthermore, with the rise in number of publicly available datasets, such as the UCR \citep{dau2019ucr} and  UEA \citep{bagnall2018uea} archives, UCI Machine Learning Repository \citep{asuncion2007uci} and Physionet Database \citep{PhysioNet}, the amount of research in the past few years has grown to quite a large extent. However, there exist two significant shortcomings in many of the existing proposed algorithms that are inherently present in deep learning.

The first shortcoming is that these algorithms assume the availability of a large amount of labeled data for building a classifier. The second shortcoming is that most machine learning and deep learning algorithms assume that the set of class labels belonging to the training and test sets are identical. In a real-world scenario, this might not be the case; say, we have a completely new class of data that we have not seen before. Unlike humans however, computer algorithms require thousands of training examples to identify a new object with a high level of accuracy. Zero shot learning (ZSL) and Generalized ZSL (GZSL) can solve the problem of identifying new classes by transferring knowledge obtained from other seen classes.


In a conventional (ZSL) setting, the set of seen classes in training data and set of unseen classes in test data are mutually disjoint. This is considered an unrealistic scenario and does not reflect standard recognition settings. Most of the time, data from the seen classes is much more abundant than that of unseen classes. It is essential to classify both seen and unseen classes at test time, rather than only unseen classes as in ZSL; such a problem is formulated as a GZSL task. In this research work, our main focus is to provide a GZSL solution for TSC.

ZSL for TSC can be used for applications such as detecting heart abnormalities from ECG data, recognizing unknown classes in human activity recognition, identifying unseen classes from sensor and spectrograph data, etc \citep{wangsensor}. ZSL and GZSL approaches have been already developed for image classification \citep{annadani2018preserving} and text \citep{alcoforado2022zeroberto} classification. To the best of our knowledge, there have not been any methods proposed to solve the problem of ZSL or GZSL for TSC.  To fill this gap, our research aims to find a solution to classify time series effectively and alleviate the major difficulties of GZSL. The key contributions of this work are summarised as follows:
\begin{enumerate}
  \item We propose a new model Latent Embedding for Time Series - GZSL (LETS-GZSL) to solve the problem of  GZSL for Time Series Classification (TSC).
  \item We  test our method on different UCR time series datasets and show that LETS-GZSL is able to achieve a harmonic mean value of at least 55\% for most of the datasets.
  \item We show that LETS-GZSL generalizes well for seen as well as unseen classes, except when the number of unseen classes are high or the amount of training data is low.
\end{enumerate}

\section{Literature Survey}
\label{LS}
In this section we present an overview of the various techniques used for time series classification, followed by various methods adopted for ZSL and GZSL. 

\subsection{Time Series Classification (TSC) Techniques}
One of the earliest and most popular approaches for TSC, often used as a strong baseline for several studies, is the use of a nearest neighbour approach coupled with a distance measure, such as the Euclidean Distance (ED) or Dynamic Time Warping (DTW) and its derivatives \citep{bagnall2017great, hsu2015flexible}. Many approaches also use time series transformations that include shapelet transforms \citep{bostrom2015binary} interval based transforms \citep{baydogan2016time}, etc. However, ensemble techniques were found to outperform their individual counterparts, which led to complex classifiers such as BOSS (Bag of SFA Symbols) \citep{schafer2015boss} and HIVE-COTE (Hierarchical Vote Collective of Transformation-based Ensembles) \citep{middlehurst2021hive}. 

Upon a surge in the popularity of deep learning, several models were proposed that were easier to train than large ensembles. Deep learning models were not only faster to train than large ensembles, but could extract complex features automatically from data. Convolutional neural networks (CNN's) worked well with time series and models such as the Time-CNN \citep{zhao2017convolutional} were proposed. Recurrent neural networks (RNN) and counterparts such as Long Short Term Networks (LSTM) \citep{pham2021time} and Echo State Networks (ESN) \citep{gallicchio2017deep} also performed well. Other approaches included autoencoder-based \citep{mehdiyev2017time}, and attention-based models \citep{ tripathi2020multivariate} 

While the aforementioned models are able to efficiently capture patterns in time series, they suffer from the constraint that most deep learning models have: they cannot generalize well to data with previously unseen class labels. In this regard, we discuss some of the existing ZSL and GZSL techniques in the following subsection.

\subsection{ZSL and GZSL Techniques}
\label{section: GZSL Lit Survey}
We present an insight into some of the existing methods of ZSL and GZSL for computer vision, the area in which most of such research has been carried out. The 'visual space' is said to comprise the images, and the 'semantic space' the semantic information, usually in the form of manually defined attributes or word vectors \citep{pourpanah2020review}. 

Several methods adopt embedding-based approaches, that try to learn a projection function from the visual space to the semantic space (and perform classification in the semantic space) \citep{chen2018zero} or vice versa \citep{annadani2018preserving}. However, such projections are often difficult to learn due to the distinctive properties of the two spaces. In this regard, latent embedding models \citep{zhang2020towards} are used which project both visual and semantic features into a latent intermediate space to explore some common properties. Other techniques such as bidirectional projections \citep{ji2020dual}, meta-learning \citep{verma2019meta}, contrastive learning \citep{han2021contrastive}, knowledge graphs \citep{bhagat2021novel} and attention mechanisms \citep{huynh2020fine} have been proposed in relation to embedding-based approaches for ZSL and GZSL.

While purely embedding based approaches are effective in the GZSL paradigm, state of the art results are often achieved with generative approaches, that generate samples for unseen classes from samples of seen classes, and semantic attributes of unseen classes under the transductive GZSL scenario i.e. when the semantic attributes of unseen classes are available during training. Generative modelling alleviates the severity of bias towards seen classes in GZSL and such models tend to have more balanced seen and unseen generalization capabilities. Several generative approaches make use of some sort of generative adversarial network (GAN) \citep{li2019leveraging}, variational autoencoder (VAE) \citep{schonfeld2019generalized} or a combination of the two \citep{narayan2020latent}. However, we follow an embedding based approach due to the inductive nature of our GZSL setting i.e. we do not have semantic information of unseen classes during training time. Additionally, generating synthetic time series that are close to the actual distribution of the unseen classes is a difficult task.


We seek to fill the void of GZSL for TSC by proposing a novel method LETS-GZSL, that makes use of latent embeddings and statistical attributes of time series, which is described in detail in section \ref{section: LETS-GZSL Methodology}.

\section{Problem Statement}
\label{PS}
In this section, we describe formally the setting of conventional ZSL as well as GZSL. We have $S$ number of seen classes in \( \mathcal{Y}_{s}\) and $U$ number of unseen classes in \( \mathcal{Y}_{u}\). The two sets are disjoint, i.e. \( \mathcal{Y}_{s} \cap \mathcal{Y}_{u} = \emptyset \). The primary goal of ZSL and GZSL is to correctly predict the class label for each time series in the test set, which we describe subsequently.

Let \( \mathcal{T} = S + U \) and \( \mathcal{Y_{T}} = \mathcal{Y}_{u} \cup \mathcal{Y}_{s} \). In both conventional and generalized ZSL, all the labelled instances used for training are from the seen classes \( \mathcal{Y}_{s}\). More formally, the training set is $ \mathcal{D}_{tr} =  \{ (x_{i}, y_{i}) |\ i = 1, 2, 3, ..., N \}$ where \(N\) is the number of time series in the training set, $x_{i}$ is a univariate time series or a set of samples at different time stamps of the form $\{ x_{i,1}, x_{i,2}, ... x_{i,T} \}$, $T$ being the number of timestamps or time series length of $x_{i}$, and $y_{i} \in \mathcal{Y}_{s}$ is the class label corresponding to the time series $x_{i}$. The fundamental difference between ZSL and GZSL lies in the test set. In conventional ZSL, the labels \(y\) come from the seen classes' set of labels only, whereas in GZSL, the labels come from both seen and unseen classes' set of labels: $\mathcal{D}_{te} = \{ (x_{i}, Y_{i}) |\ i = 1, 2, 3, ..., M \}$, where \(M\) is the number of time series in the test set, $y_{i} \in \mathcal{Y}_{u}$ for conventional ZSL and $y_{i} \in \mathcal{Y_{T}}$ for GZSL. We focus on the more realistic GZSL problem for univariate TSC in this research work.

\begin{figure*}[ht]
    \centering
    \includegraphics[scale = 0.60]{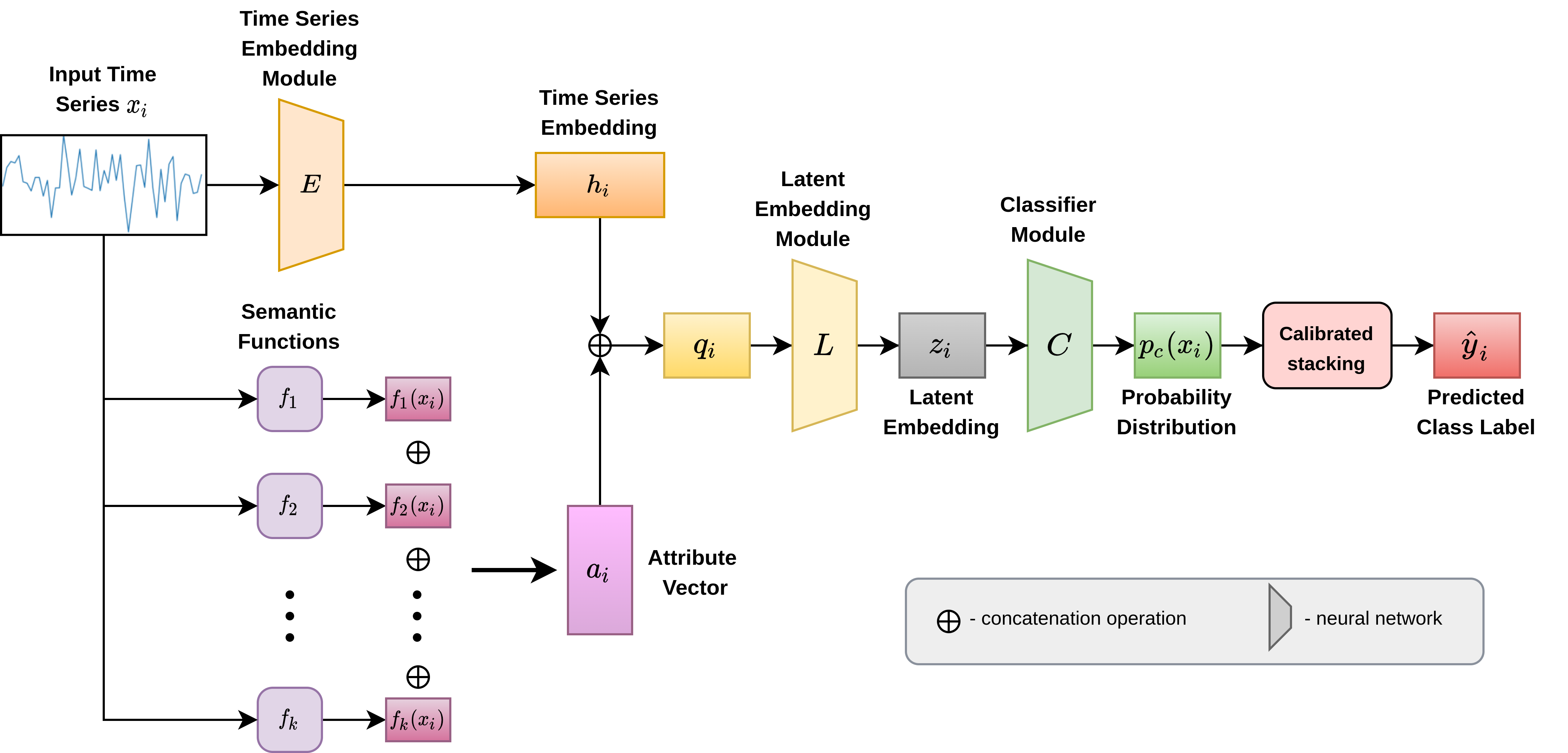}
    \caption{
    Illustration of the proposed LETS-GZSL method: We learn an embedding function \(E\) that maps each time series \(x_{i}\) into an embedding space \(h_{i} = E(x_{i})\). We also apply  several statistical functions to \(x_{i}\) whose outputs are subsequently concatenated to form the attribute vector \(a_{i}\). We then concatenate \(h_{i}\) and \(a_{i}\) and project the result into a latent space via the latent embedding module \(L\). The latent embedding \(z_{i}\) is finally passed through a classifier \(C\) and calibrated stacking is performed to get the predicted class label \(\hat{y}_{i}\)  
    }
    \label{figure:LETS-GZSL Flowchart}
\end{figure*}

\section{Our Proposed Method - Latent Embedding for Time Series GZSL}
\label{LETS}
\label{section: LETS-GZSL Methodology}
In the following section, we present our framework that uses three trained networks to effectively tackle the problem of GZSL for TSC. Figure \ref{figure:LETS-GZSL Flowchart} shows a flowchart of the various steps taken in the process. We summarize the major steps involved as follows:

\begin{enumerate}
    \item Time Series Embedding Module: We train a network with contrastive loss to obtain representative embeddings for the time series.
    \item Time Series Attribute Vectors: For every time series we obtain a vector containing statistical attributes to help discriminate between classes.
    \item Latent Embedding Module: We utilize the time series embeddings and attribute vectors from the first two steps and project them into a latent space.
    \item Classifier Module: The latent embeddings are finally fed into a neural network to predict each time series' class label.
\end{enumerate}

\subsection{Time Series Embedding Module}
\label{section: ts emb module}

\begin{figure*}[ht]
    \centering
    \includegraphics[scale=0.70]{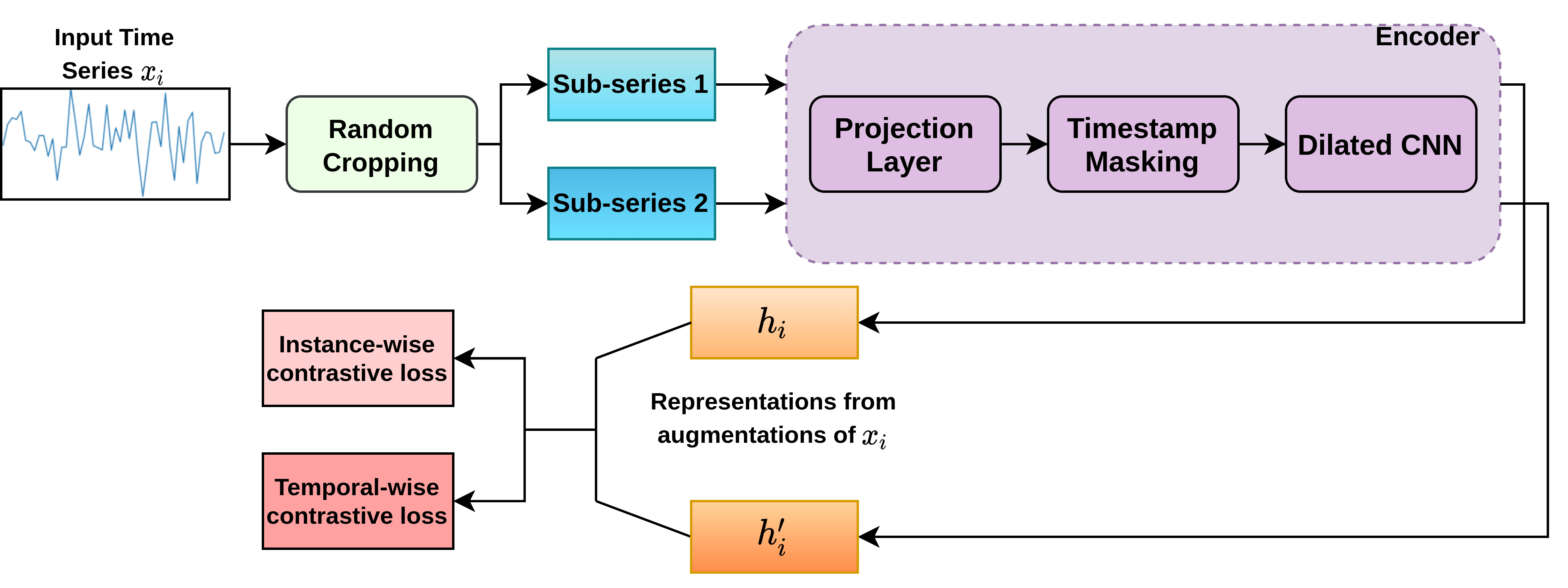}
    \caption{
    Overview of Time Series Embedding Module \(E\)
    }
    \label{figure:TS Embedding Module}
\end{figure*}

We first try to find suitable embeddings for our time series that can aid in classification. Therefore, we first learn an embedding function $E$ that can capture temporal information from the time series and extract robust contextual representations for any granularity. To do so, we adopt the method followed by \citep{yue2021ts2vec} and use contrastive losses to learn representations at various scales. 
Random cropping is first performed to augment the input time series \(x_{i}\), which generates two subseries that are subsequently passed into an encoder. The main encoder utilizes an input projection layer, timestamp masking module and dilated CNN module to obtain the representations of the two augmentations, which is depicted in Figure \ref{figure:TS Embedding Module}. Let $h_{i,t}$ and $h'_{i,t}$ denote the representations for some timestamp t, from these augmentations. Both instance wise contrastive (which views timestamps at the same position, but from different time series in a batch) and temporal wise contrastive (which views timestamps among different positions, but within the same time series) losses are used to learn the embeddings. 
The temporal contrastive loss is formulated as:
\begin{equation}
\Scale[0.9]{
\ell_{temporal}(i,t) = -\log \frac{\exp(h_{i,t} \cdot h'_{i,t})}
    {\sum_{t' \in \Lambda}(exp(h_{i,t} \cdot h'_{i,t}) + 
    \mathds{1}_{[t \neq t']}
     \exp(h_{i,t} \cdot h'_{i,t}) }
}
\end{equation}

where \(\Lambda\) is the set of timestamps within the overlap of the two subseries cropped from a given time series, and \( \mathds{1}[.] \in \{0,1\} \) is the indicator function that checks if \(t \neq t' \).

Additionally, the instance-wise contrastive loss is given by:
\begin{equation}
\Scale[0.9]{
    \ell_{instance}(i,t) = -\log \frac{exp(h_{i,t} \cdot h'_{i,t})}
    {\sum_{j = 1}^{B}(\exp(h_{i,t} \cdot h'_{j,t}) + 
    \mathds{1}_{[i \neq j]}    
    \exp(h_{i,t} \cdot h_{j,t}) }
}
\end{equation}
where \(B\) denotes the batch size.
The overall loss is given by:
\begin{equation}
    \mathcal{L}(E) = \mathbb{E}_{i,t} [\ell_{temporal}(i,t) + \ell_{instance}(i,t)]
\end{equation}

\subsection{Time Series Attribute Vectors}

\label{section: attribute vector}
Semantic information plays a crucial part in GZSL. It provides extra information about the nature of the seen and unseen classes that helps to establish the relationship between the two. In the field of computer vision, manually defined attributes or textual descriptions of images encoded as word vectors are commonly used as semantic information for performing GZSL \citep{pourpanah2020review}. However, in the case of time series, we do not have any explicitly given semantic information for either our seen or unseen classes. Instead, we introduce the use of statistical measures to derive an attribute vector \( a_{i} \in \mathbb{R}^{K} \) for a corresponding time series \(x_{i}\), that could further help discriminate between the seen and unseen classes. Thus, every attribute vector can be represented as: 
\begin{equation}
    \vect{a_{i}} = [\ f_{1}(x_{i}) \; f_{2}(x_{i}) \; f_{3}(x_{i}) \;  ... \;  f_{K}(x_{i})\ ] 
\end{equation}
where
\( f_{j} \;  , j=1,2,3, ... , K \) are different functions applied on \( x_{i} \) and
$K$ is the number of such mathematical functions.

\subsection{Latent Embedding Module}
Once the time series embeddings and attribute vectors are obtained as specified in sections \ref{section: ts emb module} and \ref{section: attribute vector}, they are jointly projected into a latent space. This latent space is the final space in which the classification is performed. We call this module $L$ that learns the latent embedding as the classifier module. The input $q_{i}$ to $L$ consists of the concatenation of a time series' embedding $h_{i}$ and its respective attribute vector $a_{i}$, i.e. \(q_{i} = h_{i}  \oplus  a_{i}\).
A convolutional neural network is trained and the final latent embedding \(z_{i} = L(q_{i})\) , \( z_{i} \in \mathbb{R}^{P} \) is learned, where P is the dimension of the latent space.

\subsection{Classifier Module}

\begin{figure*}[t]
    \centering
    \includegraphics[scale=0.70]{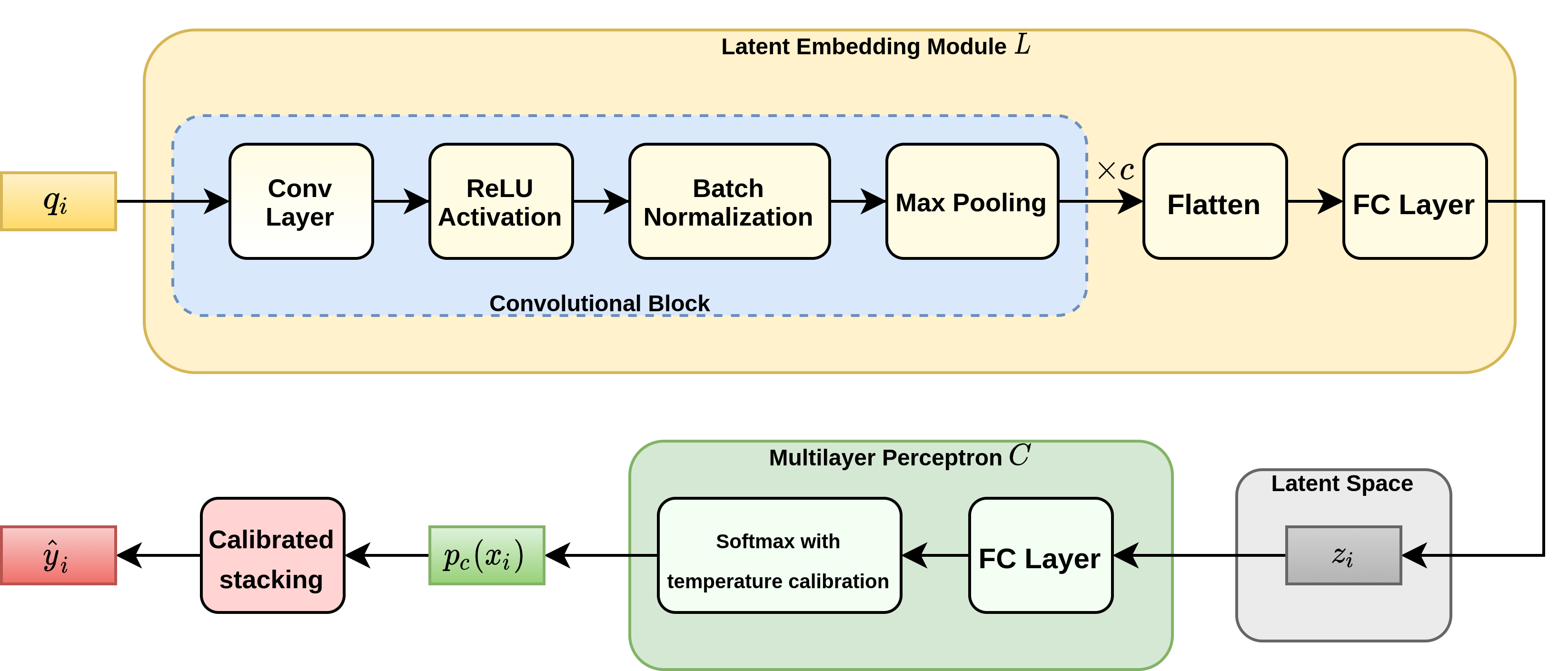}
    \caption{Architecture of the Latent Embedding Module and Classifier Module}
    \label{figure: L and C Module}
\end{figure*}

The latent embeddings are finally fed through a feedforward neural network $C$ that outputs a vector of logits \( r = C(z_{i}) \) for a given time series. The final probability distribution for a given time series $x_{i}$ over the $S$ seen classes is given as follows:
\begin{equation}
\label{equation: seen_prob_dist}
    p_{c}(x_{i}) = \frac{\exp(r_{c} / \tau)}
                        {\sum_{c'=1}^{S} \exp(r_{c'} / \tau)}
\end{equation}
where $\tau$ is the temperature parameter used in temperature calibration \citep{hinton2015distilling}, which discourages the model from generating overconfident probabilities, which is especially useful since seen class probabilities tend to be much higher than unseen ones. Additionally, we also use calibrated stacking \citep{chao2016empirical} which intuitively reduces the scores of the seen classes. The final predicted class label \(\hat{y}_{i}\) is given as: 

\begin{equation}
\label{equation: calibrated stacking}
    \hat{y}_{i} = \argmax_{c \in \mathcal{Y_{T}}} \ 
                p_{c}(x_{i}) - \gamma \mathds{1}[c \in \mathcal{Y}_{s}]
\end{equation}

where $\gamma$ is a hyperparameter known as the calibration factor and $\mathds{1}[.] = 1$ if \(c \in \mathcal{Y}_{s}\) and 0 otherwise. An optimal combination of the hyperparameters $\tau$ and $\gamma$ can penalize high seen class probabilities and lead to effective classification of unseen test samples as well.

The training loss for learning networks L and C is the cross-entropy between the probabilities and true labels (in the form of one-hot vectors), given by:
\begin{equation}
\label{equation: classifier_loss}
    \mathcal{L}(C, L) = - \sum_{i=1}^{N} \sum_{c=1}^{S} y_{i,c} \ \log p_{c}(x_{i})
\end{equation}

The overall architecture of the \(L\) and \(C\) is depicted in Figure \ref{figure: L and C Module}.

\section{Experimental Details}
\label{ED}
\subsection{Datasets}
\label{section: datasets}
To test out the effectiveness of our proposed method, we use datasets from the UCR Archive \citep{dau2019ucr}, which contains several different datasets for univariate time series classification. We choose datasets from different domains that differ in properties such as dataset size (the number of time series in the dataset), number of classes and time series length to study how LETS-GZSL performs with such variations. The datasets used and their properties can be found in Table \ref{table: dataset description}:

\begin{table}[h]
\centering
\caption{UCR Archive Datasets' Characteristics}
\label{table: dataset description}
\resizebox{\linewidth}{!}{%
\begin{tabular}{|c|c|c|c|c|}
\hline
Dataset              & Type      & Size  & \begin{tabular}[c]{@{}c@{}}Number \\ of Classes\end{tabular} & \multicolumn{1}{l|}{Length} \\ \hline
TwoPatterns          & Simulated & 5000  & 4                                                            & 128                         \\
ElectricDevices      & Device    & 16637 & 7                                                            & 96                          \\
Trace                & Sensor    & 200   & 4                                                            & 275                         \\
SyntheticControl     & Simulated & 600   & 6                                                            & 60                          \\
UWaveGestureLibraryX & Motion    & 4478  & 8                                                            & 945                         \\
CricketX             & Motion    & 780   & 12                                                           & 300                         \\
Beef                 & Spectro   & 60    & 5                                                            & 470                         \\
InsectWingbeatSound  & Sensor    & 2200  & 11                                                           & 256                         \\ \hline
\end{tabular}}
\end{table}

\subsection{Validation Scheme}
\label{section: validation scheme}
Whereas in classical machine learning and deep learning, datasets are divided into train, validation and test sets sample-wise, in ZSL and GZSL, they are divided class-wise. Originally, we keep \(\lceil \frac{2U}{3} \rceil\) classes as training and validation classes and the rest as test classes. 20\% of the total samples belonging to the training and validation classes are kept aside as the seen test set. The samples from the test classes are kept as the unseen test set. 
To arrive at the optimal hyperparameters, we require a validation set, which we create by splitting the \(\lceil \frac{2U}{3} \rceil\) classes again, same as before. Half the classes are kept for training and the other half for validation. 20\% of the total samples belonging to the training classes are kept aside as the seen validation set, and the remaining as the unseen validation set. The division of classes is chosen so as to keep the number of samples in the validation and the test sets roughly the same. 

\begin{figure}[h]
    \centering
    \includegraphics[scale=0.70]{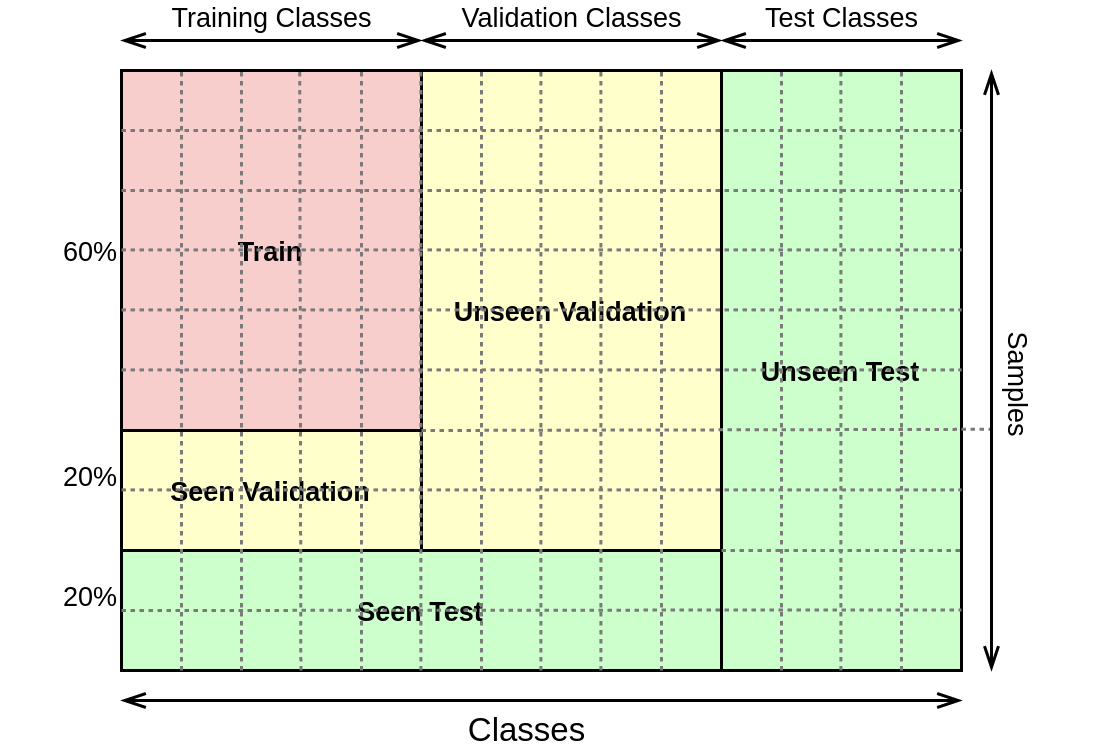}
    \caption{Validation Scheme: Each row represents number of samples and column represent number of classes. Note that the number of samples in the three sets may be different.}
    \label{figure: validation scheme}
\end{figure}

The validation scheme \citep{le2019classical}, is further shown diagrammatically in Figure 
\ref{figure: validation scheme} for a better understanding. Essentially, we are given the samples as shown in red and yellow during training time, for which we need to split appropriately into training and validation sets, to correctly predict the samples as shown in green during test time. Therefore, the LETS-GZSL model is first trained on the training set, and hyperparameters are tuned based on the metrics evaluated on the seen validation and unseen validation sets. The model is finally retrained on the train, seen and unseen validation sets, and the final results are reported on the seen and unseen test sets. 

\subsection{Evaluation Metrics}
Under the GZSL scenario, we measure the top-1 accuracy on the seen and unseen classes, denoted as $acc_{s}$ and $acc_{u}$ respectively. Another metric we measure is the widely popular harmonic mean, which is able to measure the inherent bias towards seen classes and is defined as follows:
\begin{equation}
    H = \frac{2 \times acc_{s} \times acc_{u}}{acc_{s} + acc_{u}}
\end{equation}

We also report the area under seen unseen curve (AUSUC), introduced by \citep{chao2016empirical}. There is an intrinsic tradeoff between the seen and unseen accuracy of a GZSL model and the AUSUC metric measures the tradeoff between the two. The AUSUC metric is calculated by varying the value of $\gamma$ in equation \ref{equation: calibrated stacking}, noting the seen and unseen accuracies and then calculating the area under the curve given by the two quantities. Models with higher values of AUSUC achieve better balanced performance.

\subsection{Implementation Details}

We first find out the best hyperparameters for our embedding module for a given dataset. The main hyperparameters for $E$ include the output dimensions of the representative embeddings, hidden dimension of the encoder, number of residual blocks in the encoder and batch size for training. The batch size is set to a larger value of 128 as suggested by \citet{yue2021ts2vec} to incorporate enough negative samples for calculating the instance-wise contrastive loss. The remaining hyperparameters are decided via a random search. We use an Adam optimizer \citep{kingma2014adam} with $10^{-3}$ as the initial learning rate to arrive at the optimal weights.

To compute the attribute vector $a_{i}$ for each time series, we use statistical measures that include mean, median, max, argmax, min, argmin, skew, kurtosis and approximate entropy.

The latent embedding module $L$ consists of convolutional blocks which is subsequently flattened and fed into a fully connected (FC) layer of the required latent embedding dimension $P$. The latent dimensions, number of convolutional blocks (denoted by \(c\)), filter sizes, number of filters, pool size and other hyperparameters of $L$ are tuned for each dataset separately due to the varying lengths and nature of the time series. The hyperparameters of the classifier module $C$ are also tuned simultaneously. We utilize the Tree of Parzen Estimators algorithm \citep{bergstra2011algorithms} while using AUSUC as the metric to choose the optimal hyperparameters. 

\section{Results and Discussion}
\label{RD}
We evaluate the LETS-GZSL method on the datasets mentioned in section \ref{section: datasets} and discuss the key results and observations. The various evaluation metrics recorded across the datasets are tabulated in table \ref{table: evaluation metrics}. To obtain the AUSUC values and plots, we vary \(\gamma\) from -1 to 0 using an interval of 0.05, then vary \(\gamma\) from 0 to 1, with an interval of 0.001, noting down the seen and unseen accuracies for each value.

\begin{table}[h]
\centering
\caption{Evaluation metrics of LETS-GZSL on various UCR datasets}
\label{table: evaluation metrics}
\resizebox{\linewidth}{!}{%
\begin{tabular}{|c|c|c|c|c|}
\hline
Dataset              & AUSUC   & \(acc_{s}\) & \(acc_{u}\) & \(H\)  \\ \hline
TwoPatterns          & 95.233  & 94.504      & 88.815      & 91.571 \\
ElectricDevices      & 55.749  & 60.971      & 55.323      & 58.010 \\
Trace                & 52.699  & 43.333      & 76.000      & 55.195 \\
SyntheticControl     & 46.028  & 50.000      & 68.500      & 57.805 \\
UWaveGestureLibraryX & 44.671  & 58.944      & 51.636      & 55.049 \\
CricketX             & 27.328  & 50.427      & 39.487      & 44.291 \\
Beef                 & 25.000  & 10.000      & 100.000     & 18.181 \\
InsectWingbeatSound  & 18.907  & 30.625      & 35.500      & 32.882 \\ \hline
\end{tabular}}
\end{table}

\begin{figure}[h]
    \centering
    \includegraphics[scale=0.25]{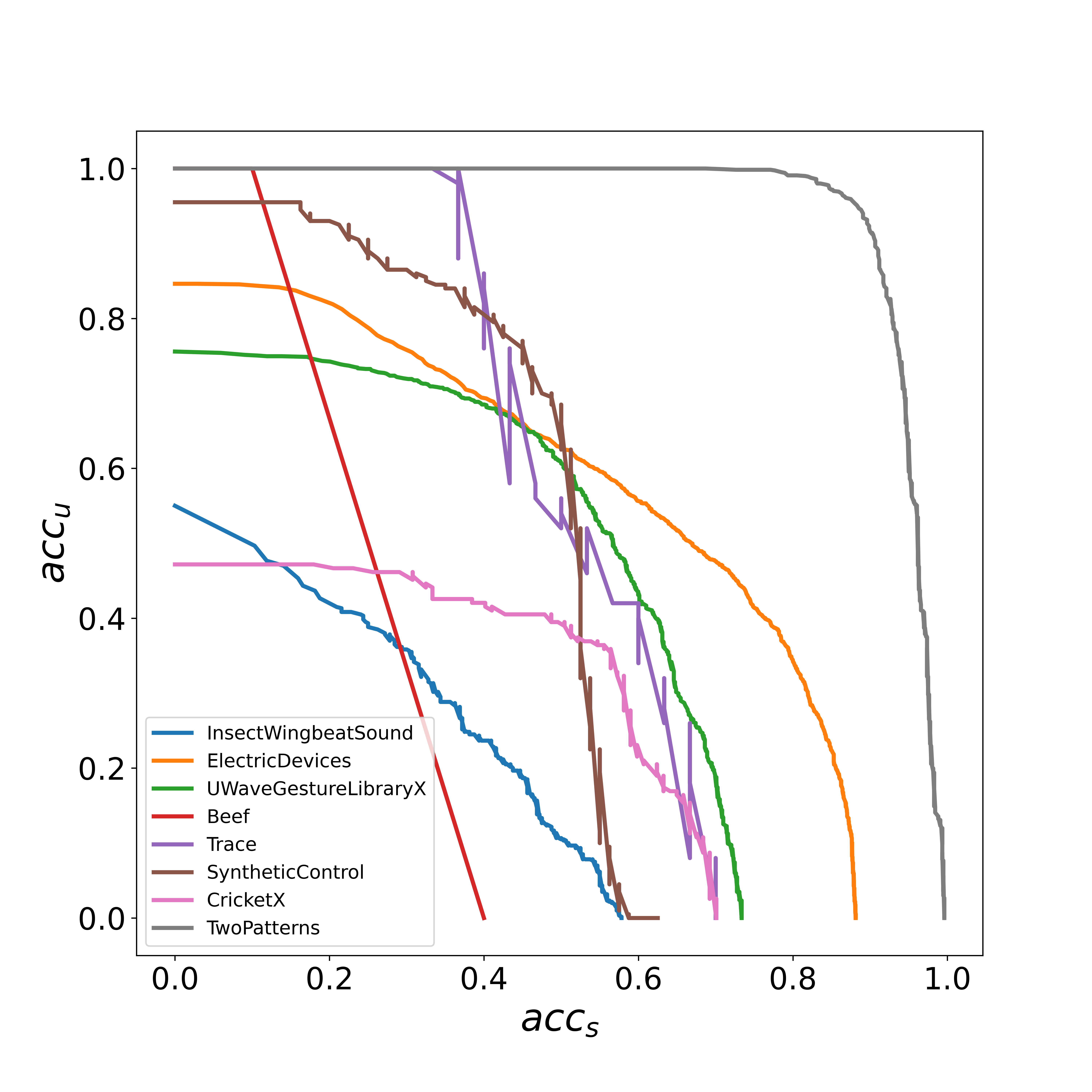}
    \caption{AUSUC curves for LETS-GZSL evaluation on various datasets}
    \label{figure: AUSUC Plot}
\end{figure}

From the AUSUC values, we notice that LETS-GZSL is able to achieve good accuracies with both seen and unseen classes on most of the datasets. However, there are two factors that greatly influence the AUSUC. We discuss them below and provide plausible explanations for their effects.

The first factor is the number of classes, which has an inverse relationship to the AUSUC. Noticeably, although the harmonic mean is relatively high for datasets even with a large number of classes, the AUSUC is low in such cases, as can be seen in the CricketX and InsectWingbeatSound datasets. This implies that for some value of \(\gamma\) and \(\tau\) LETS-GZSL can efficiently decrease the bias towards seen classes, yet lacks the intrinsic ability to balance the seen and unseen accuracy. We attribute this to the fact that our model can learn from seen class examples easily, but has no 'efficient mechanism' to distinguish between unseen class examples themselves. This is further demonstrated by the fact that calibrated stacking penalizes all unseen class scores by an equal amount. 

The second factor is the amount of data, i.e. size of the dataset. Scarcity of labelled data has always posed a problem for deep learning models and we concur that the same problem applies to our framework. LETS-GZSL is not able to generalize to new seen class samples by simply seeing a couple of such samples during training. In this scenario, the predictions are almost equal to random guesses. Therefore, we notice from Figure \ref{figure: AUSUC Plot} that the AUSUC curves of smaller datasets are noisier as compared to larger ones. From Table \ref{table: evaluation metrics} it is evident that the TwoPatterns dataset, which is 25 times larger than the Trace dataset, has a much higher AUSUC despite the two having the same number of classes. Although the Beef dataset has only 5 classes, with 2 for training and validation each and one for testing, due to its extremely small size, the AUSUC and \(acc_{s}\) values are very low. 

We also notice another interesting observation in the values of the harmonic mean for the various datasets. For datasets with smaller number of classes or limited data, the maximum value of \(H\) is achieved when the seen class scores are penalized heavily and \(acc_{s} < acc_{u}\). For most datasets with relatively larger number of classes, \(acc_{s}\) tends to be greater than \(acc_{u}\). For example, in the Beef dataset, \(acc_{s} = 10\%\), while \(acc_{u} = 100\%\). This could be due to the insufficient amount of data for recognizing seen classes themselves, as well as the ability to classify all the unseen class test examples correctly by simply using a large value of \(\gamma\), since there exists only one unseen test class.

We further conduct ablation studies that can be found in the supplementary section to study how each component affects the model performance. Additionally, we also study performance against a deep learning baseline: the Long Short Term Memory-Fully Convolutional Network (LSTM-FCN) \citep{karim2019insights}, which shows the superiority of our model.

\section{Conclusions and Future Work}
\label{section: conclusion and future work}
In this paper, we have proposed the use of a latent embedding model for the purpose of GZSL for TSC. The proposed LETS-GZSL model uses contrastive loss to learn embeddings for time series data and projects it simultaneously along with statistical attributes to a latent space. A classifier is trained on these latent embeddings, with temperature calibration and calibrated stacking to aid in classification. We have shown that LETS-GZSL is able to generalize well to both seen and unseen class example, except when the number of unseen classes are high or the amount of training data is low.

In the future, techniques to solve the GZSL problem for TSC in the above mentioned cases of large number of unseen classes or limited amount of seen class data, can be explored. In this direction, meta learning methods and generative approaches can be employed to offset the above problem as well as class imbalance problems. Such techniques for multivariate TSC are also yet to be experimented with.

\bibliographystyle{named}
\bibliography{references}

\begin{thebibliography}{}

\bibitem[\protect\citeauthoryear{Abanda \bgroup \em et al.\egroup
  }{2019}]{abanda2019review}
Amaia Abanda, Usue Mori, and Jose~A Lozano.
\newblock A review on distance based time series classification.
\newblock {\em Data Mining and Knowledge Discovery}, 33(2):378--412, 2019.

\bibitem[\protect\citeauthoryear{Alcoforado \bgroup \em et al.\egroup
  }{2022}]{alcoforado2022zeroberto}
Alexandre Alcoforado, Thomas~Palmeira Ferraz, Rodrigo Gerber, Enzo Bustos,
  Andr{\'e}~Seidel Oliveira, Bruno~Miguel Veloso, Fabio~Levy Siqueira, and Anna
  Helena~Reali Costa.
\newblock Zeroberto--leveraging zero-shot text classification by topic
  modeling.
\newblock {\em arXiv preprint arXiv:2201.01337}, 2022.

\bibitem[\protect\citeauthoryear{Annadani and
  Biswas}{2018}]{annadani2018preserving}
Yashas Annadani and Soma Biswas.
\newblock Preserving semantic relations for zero-shot learning.
\newblock In {\em Proceedings of the IEEE Conference on Computer Vision and
  Pattern Recognition}, pages 7603--7612, 2018.

\bibitem[\protect\citeauthoryear{Asuncion and Newman}{2007}]{asuncion2007uci}
Arthur Asuncion and David Newman.
\newblock Uci machine learning repository, 2007.

\bibitem[\protect\citeauthoryear{Bagnall \bgroup \em et al.\egroup
  }{2017}]{bagnall2017great}
Anthony Bagnall, Jason Lines, Aaron Bostrom, James Large, and Eamonn Keogh.
\newblock The great time series classification bake off: a review and
  experimental evaluation of recent algorithmic advances.
\newblock {\em Data mining and knowledge discovery}, 31(3):606--660, 2017.

\bibitem[\protect\citeauthoryear{Bagnall \bgroup \em et al.\egroup
  }{2018}]{bagnall2018uea}
Anthony Bagnall, Hoang~Anh Dau, Jason Lines, Michael Flynn, James Large, Aaron
  Bostrom, Paul Southam, and Eamonn Keogh.
\newblock The uea multivariate time series classification archive, 2018.
\newblock {\em arXiv preprint arXiv:1811.00075}, 2018.

\bibitem[\protect\citeauthoryear{Baydogan and Runger}{2016}]{baydogan2016time}
Mustafa~Gokce Baydogan and George Runger.
\newblock Time series representation and similarity based on local
  autopatterns.
\newblock {\em Data Mining and Knowledge Discovery}, 30(2):476--509, 2016.

\bibitem[\protect\citeauthoryear{Bergstra \bgroup \em et al.\egroup
  }{2011}]{bergstra2011algorithms}
James Bergstra, R{\'e}mi Bardenet, Yoshua Bengio, and Bal{\'a}zs K{\'e}gl.
\newblock Algorithms for hyper-parameter optimization.
\newblock {\em Advances in neural information processing systems}, 24, 2011.

\bibitem[\protect\citeauthoryear{Bhagat \bgroup \em et al.\egroup
  }{2021}]{bhagat2021novel}
PK~Bhagat, Prakash Choudhary, Kh~Singh, et~al.
\newblock A novel approach based on fully connected weighted bipartite graph
  for zero-shot learning problems.
\newblock {\em Journal of Ambient Intelligence and Humanized Computing},
  12(9):8647--8662, 2021.

\bibitem[\protect\citeauthoryear{Bostrom and Bagnall}{2015}]{bostrom2015binary}
Aaron Bostrom and Anthony Bagnall.
\newblock Binary shapelet transform for multiclass time series classification.
\newblock In {\em International conference on big data analytics and knowledge
  discovery}, pages 257--269. Springer, 2015.

\bibitem[\protect\citeauthoryear{Chao \bgroup \em et al.\egroup
  }{2016}]{chao2016empirical}
Wei-Lun Chao, Soravit Changpinyo, Boqing Gong, and Fei Sha.
\newblock An empirical study and analysis of generalized zero-shot learning for
  object recognition in the wild.
\newblock In {\em European conference on computer vision}, pages 52--68.
  Springer, 2016.

\bibitem[\protect\citeauthoryear{Chen \bgroup \em et al.\egroup
  }{2018}]{chen2018zero}
Long Chen, Hanwang Zhang, Jun Xiao, Wei Liu, and Shih-Fu Chang.
\newblock Zero-shot visual recognition using semantics-preserving adversarial
  embedding networks.
\newblock In {\em Proceedings of the IEEE conference on computer vision and
  pattern recognition}, pages 1043--1052, 2018.

\bibitem[\protect\citeauthoryear{Dau \bgroup \em et al.\egroup
  }{2019}]{dau2019ucr}
Hoang~Anh Dau, Anthony Bagnall, Kaveh Kamgar, Chin-Chia~Michael Yeh, Yan Zhu,
  Shaghayegh Gharghabi, Chotirat~Ann Ratanamahatana, and Eamonn Keogh.
\newblock The ucr time series archive.
\newblock {\em IEEE/CAA Journal of Automatica Sinica}, 6(6):1293--1305, 2019.

\bibitem[\protect\citeauthoryear{Esmael \bgroup \em et al.\egroup
  }{2012}]{esmael2012improving}
Bilal Esmael, Arghad Arnaout, Rudolf~K Fruhwirth, and Gerhard Thonhauser.
\newblock Improving time series classification using hidden markov models.
\newblock In {\em 2012 12th International Conference on Hybrid Intelligent
  Systems (HIS)}, pages 502--507. IEEE, 2012.

\bibitem[\protect\citeauthoryear{Fawaz \bgroup \em et al.\egroup
  }{2019}]{fawaz2019deep}
Hassan~Ismail Fawaz, Germain Forestier, Jonathan Weber, Lhassane Idoumghar, and
  Pierre-Alain Muller.
\newblock Deep learning for time series classification: a review.
\newblock {\em Data mining and knowledge discovery}, 33(4):917--963, 2019.

\bibitem[\protect\citeauthoryear{Gallicchio and
  Micheli}{2017}]{gallicchio2017deep}
Claudio Gallicchio and Alessio Micheli.
\newblock Deep echo state network (deepesn): A brief survey.
\newblock {\em arXiv preprint arXiv:1712.04323}, 2017.

\bibitem[\protect\citeauthoryear{Goldberger \bgroup \em et al.\egroup
  }{2000}]{PhysioNet}
A.~L. Goldberger, L.~A.~N. Amaral, L.~Glass, J.~M. Hausdorff, P.~Ch. Ivanov,
  R.~G. Mark, J.~E. Mietus, G.~B. Moody, C.-K. Peng, and H.~E. Stanley.
\newblock {PhysioBank, PhysioToolkit, and PhysioNet}: Components of a new
  research resource for complex physiologic signals.
\newblock {\em Circulation}, 101(23):e215--e220, 2000.
\newblock Circulation Electronic Pages:
  http://circ.ahajournals.org/content/101/23/e215.full PMID:1085218; doi:
  10.1161/01.CIR.101.23.e215.

\bibitem[\protect\citeauthoryear{Han \bgroup \em et al.\egroup
  }{2021}]{han2021contrastive}
Zongyan Han, Zhenyong Fu, Shuo Chen, and Jian Yang.
\newblock Contrastive embedding for generalized zero-shot learning.
\newblock In {\em Proceedings of the IEEE/CVF Conference on Computer Vision and
  Pattern Recognition}, pages 2371--2381, 2021.

\bibitem[\protect\citeauthoryear{Hinton \bgroup \em et al.\egroup
  }{2015}]{hinton2015distilling}
Geoffrey Hinton, Oriol Vinyals, and Jeff Dean.
\newblock Distilling the knowledge in a neural network.
\newblock {\em arXiv preprint arXiv:1503.02531}, 2015.

\bibitem[\protect\citeauthoryear{Hsu \bgroup \em et al.\egroup
  }{2015}]{hsu2015flexible}
Che-Jui Hsu, Kuo-Si Huang, Chang-Biau Yang, and Yi-Pu Guo.
\newblock Flexible dynamic time warping for time series classification.
\newblock {\em Procedia Computer Science}, 51:2838--2842, 2015.

\bibitem[\protect\citeauthoryear{Huynh and Elhamifar}{2020}]{huynh2020fine}
Dat Huynh and Ehsan Elhamifar.
\newblock Fine-grained generalized zero-shot learning via dense attribute-based
  attention.
\newblock In {\em Proceedings of the IEEE/CVF conference on computer vision and
  pattern recognition}, pages 4483--4493, 2020.

\bibitem[\protect\citeauthoryear{Ji \bgroup \em et al.\egroup
  }{2020}]{ji2020dual}
Zhong Ji, Hai Wang, Yanwei Pang, and Ling Shao.
\newblock Dual triplet network for image zero-shot learning.
\newblock {\em Neurocomputing}, 373:90--97, 2020.

\bibitem[\protect\citeauthoryear{Karim \bgroup \em et al.\egroup
  }{2019}]{karim2019insights}
Fazle Karim, Somshubra Majumdar, and Houshang Darabi.
\newblock Insights into lstm fully convolutional networks for time series
  classification.
\newblock {\em IEEE Access}, 7:67718--67725, 2019.

\bibitem[\protect\citeauthoryear{Kingma and Ba}{2014}]{kingma2014adam}
Diederik~P Kingma and Jimmy Ba.
\newblock Adam: A method for stochastic optimization.
\newblock {\em arXiv preprint arXiv:1412.6980}, 2014.

\bibitem[\protect\citeauthoryear{Le~Cacheux \bgroup \em et al.\egroup
  }{2019}]{le2019classical}
Yannick Le~Cacheux, Herv{\'e} Le~Borgne, and Michel Crucianu.
\newblock From classical to generalized zero-shot learning: A simple adaptation
  process.
\newblock In {\em International Conference on Multimedia Modeling}, pages
  465--477. Springer, 2019.

\bibitem[\protect\citeauthoryear{Lei and Wu}{2020}]{lei2020time}
Yuxia Lei and Zhongqiang Wu.
\newblock Time series classification based on statistical features.
\newblock {\em EURASIP Journal on Wireless Communications and Networking},
  2020(1):1--13, 2020.

\bibitem[\protect\citeauthoryear{Li \bgroup \em et al.\egroup
  }{2019}]{li2019leveraging}
Jingjing Li, Mengmeng Jing, Ke~Lu, Zhengming Ding, Lei Zhu, and Zi~Huang.
\newblock Leveraging the invariant side of generative zero-shot learning.
\newblock In {\em Proceedings of the IEEE/CVF Conference on Computer Vision and
  Pattern Recognition}, pages 7402--7411, 2019.

\bibitem[\protect\citeauthoryear{Mehdiyev \bgroup \em et al.\egroup
  }{2017}]{mehdiyev2017time}
Nijat Mehdiyev, Johannes Lahann, Andreas Emrich, David Enke, Peter Fettke, and
  Peter Loos.
\newblock Time series classification using deep learning for process planning:
  A case from the process industry.
\newblock {\em Procedia Computer Science}, 114:242--249, 2017.

\bibitem[\protect\citeauthoryear{Middlehurst \bgroup \em et al.\egroup
  }{2021}]{middlehurst2021hive}
Matthew Middlehurst, James Large, Michael Flynn, Jason Lines, Aaron Bostrom,
  and Anthony Bagnall.
\newblock Hive-cote 2.0: a new meta ensemble for time series classification.
\newblock {\em arXiv preprint arXiv:2104.07551}, 2021.

\bibitem[\protect\citeauthoryear{Narayan \bgroup \em et al.\egroup
  }{2020}]{narayan2020latent}
Sanath Narayan, Akshita Gupta, Fahad~Shahbaz Khan, Cees~GM Snoek, and Ling
  Shao.
\newblock Latent embedding feedback and discriminative features for zero-shot
  classification.
\newblock In {\em European Conference on Computer Vision}, pages 479--495.
  Springer, 2020.

\bibitem[\protect\citeauthoryear{Pham}{2021}]{pham2021time}
Tuan~D Pham.
\newblock Time--frequency time--space lstm for robust classification of
  physiological signals.
\newblock {\em Scientific reports}, 11(1):1--11, 2021.

\bibitem[\protect\citeauthoryear{Pourpanah \bgroup \em et al.\egroup
  }{2020}]{pourpanah2020review}
Farhad Pourpanah, Moloud Abdar, Yuxuan Luo, Xinlei Zhou, Ran Wang, Chee~Peng
  Lim, and Xi-Zhao Wang.
\newblock A review of generalized zero-shot learning methods.
\newblock {\em arXiv preprint arXiv:2011.08641}, 2020.

\bibitem[\protect\citeauthoryear{Sch{\"a}fer}{2015}]{schafer2015boss}
Patrick Sch{\"a}fer.
\newblock The boss is concerned with time series classification in the presence
  of noise.
\newblock {\em Data Mining and Knowledge Discovery}, 29(6):1505--1530, 2015.

\bibitem[\protect\citeauthoryear{Schonfeld \bgroup \em et al.\egroup
  }{2019}]{schonfeld2019generalized}
Edgar Schonfeld, Sayna Ebrahimi, Samarth Sinha, Trevor Darrell, and Zeynep
  Akata.
\newblock Generalized zero-and few-shot learning via aligned variational
  autoencoders.
\newblock In {\em Proceedings of the IEEE/CVF Conference on Computer Vision and
  Pattern Recognition}, pages 8247--8255, 2019.

\bibitem[\protect\citeauthoryear{Tripathi and
  Baruah}{2020}]{tripathi2020multivariate}
Achyut~Mani Tripathi and Rashmi~Dutta Baruah.
\newblock Multivariate time series classification with an attention-based
  multivariate convolutional neural network.
\newblock In {\em 2020 International Joint Conference on Neural Networks
  (IJCNN)}, pages 1--8. IEEE, 2020.

\bibitem[\protect\citeauthoryear{Verma \bgroup \em et al.\egroup
  }{2019}]{verma2019meta}
Vinay~Kumar Verma, Dhanajit Brahma, and Piyush Rai.
\newblock A meta-learning framework for generalized zero-shot learning.
\newblock {\em arXiv preprint arXiv:1909.04344}, 2019.

\bibitem[\protect\citeauthoryear{Wang}{}]{wangsensor}
Wei Wang.
\newblock Sensor-based human activity recognition via zero-shot learning.

\bibitem[\protect\citeauthoryear{Wang \bgroup \em et al.\egroup
  }{2006}]{wang2006characteristic}
Xiaozhe Wang, Kate Smith, and Rob Hyndman.
\newblock Characteristic-based clustering for time series data.
\newblock {\em Data mining and knowledge Discovery}, 13(3):335--364, 2006.

\bibitem[\protect\citeauthoryear{Yu \bgroup \em et al.\egroup
  }{2021}]{yu2021analysis}
Wennian Yu, Il~Yong Kim, and Chris Mechefske.
\newblock Analysis of different rnn autoencoder variants for time series
  classification and machine prognostics.
\newblock {\em Mechanical Systems and Signal Processing}, 149:107322, 2021.

\bibitem[\protect\citeauthoryear{Yue \bgroup \em et al.\egroup
  }{2021}]{yue2021ts2vec}
Zhihan Yue, Yujing Wang, Juanyong Duan, Tianmeng Yang, Congrui Huang, Yunhai
  Tong, and Bixiong Xu.
\newblock Ts2vec: Towards universal representation of time series.
\newblock {\em arXiv preprint arXiv:2106.10466}, 2021.

\bibitem[\protect\citeauthoryear{Zhang \bgroup \em et al.\egroup
  }{2020}]{zhang2020towards}
Lei Zhang, Peng Wang, Lingqiao Liu, Chunhua Shen, Wei Wei, Yanning Zhang, and
  Anton Van Den~Hengel.
\newblock Towards effective deep embedding for zero-shot learning.
\newblock {\em IEEE Transactions on Circuits and Systems for Video Technology},
  30(9):2843--2852, 2020.

\bibitem[\protect\citeauthoryear{Zhao \bgroup \em et al.\egroup
  }{2017}]{zhao2017convolutional}
Bendong Zhao, Huanzhang Lu, Shangfeng Chen, Junliang Liu, and Dongya Wu.
\newblock Convolutional neural networks for time series classification.
\newblock {\em Journal of Systems Engineering and Electronics}, 28(1):162--169,
  2017.

\end{thebibliography}

\newpage

\appendix
\section{Ablation Studies}
\label{section: ablation studies}

We perform ablation studies to measure the impact of the LETS-GZSL components on the performance of the overall framework. More specifically, we remove the embedding module \(E\) and attribute vectors of the time series \(a_i\) one at a time and record the metrics on the datasets. We also give plausible reasons for the observations.

\begin{table}[h]
\caption{Results without Embedding Module E}
\label{table: ablation E}
\resizebox{\linewidth}{!}{%
\begin{tabular}{|c|c|c|c|c|}
\hline
Dataset Name         & AUSUC  & $acc_s$ & $acc_u$ & $H$      \\ \hline
TwoPatterns          & 80.628 & 84.455 & 73.290 & 81.791 \\
ElectricDevices      & 53.768 & 54.904 & 48.977 & 51.771 \\
Trace                & 49.933 & 50.613 & 41.526 & 45.621 \\
Synthetic Control    & 44.102 & 65.004 & 47.722 & 55.038 \\
UWaveGestureLibraryX & 42.051 & 56.809 & 49.134 & 52.693 \\
CricketX             & 25.447 & 48.560 & 37.334 & 42.213 \\
Beef                 & 14.216 & 12.125 & 20.375 & 15.202 \\
InsectWingbeatSound  & 18.565 & 28.725 & 27.036 & 27.854 \\ \hline
\end{tabular}}
\end{table}

\begin{table}[h]
\caption{Results without Stastical Attributes}
\label{table: ablation A}
\resizebox{\linewidth}{!}{%
\begin{tabular}{|c|c|c|c|c|}
\hline
Dataset Name         & AUSUC  & $acc_s$ & $acc_u$ & $H$      \\ \hline
TwoPatterns          & 93.136 & 93.852 & 86.908 & 90.246 \\
ElectricDevices      & 54.343 & 58.075 & 53.162 & 55.510 \\
Trace                & 50.664 & 45.734 & 72.153 & 55.983 \\
Synthetic Control    & 45.093 & 49.607 & 65.398 & 56.418 \\
UWaveGestureLibraryX & 43.589 & 57.216 & 50.574 & 53.69  \\
CricketX             & 26.147 & 47.997 & 38.353 & 42.636 \\
Beef                 & 20.702 & 8.750  & 80.000 & 15.774 \\
InsectWingbeatSound  & 17.841 & 29.119 & 32.675 & 30.790 \\ \hline
\end{tabular}}
\end{table}

On observing the results of the ablation studies we notice that the performance is not as good as that of the entire LETS-GZSL model. 

In our first experiment, we remove the embedding module \(E\) and instead concatenate the raw input time series with the statistical features into the classifier module. From table \ref{table: ablation E} we notice that there is a major drop in performance in doing so. All the metrics, i.e., AUSUC, seen accuracy, unseen accuracy and harmonic mean are significantly lesser. This is possibly due to the inability of our model to discriminate between seen and unseen classes with raw input data, i.e. the time series. The embedding module has extracted meaningful representative features from the time series for the classifier module to discriminate between the classes. Additonally, the seen accuracy is greater than the seen accuracy for almost all the datasets.

In our second experiment, we utilize the embedding module \(E\) but do not concatenate any statistical features. The time series representations are directly fed into the classifier module. Here we notice that there is a much smaller dropoff in performance. For most datasets, this dropoff is around by 1 or 2 units in AUSUC. Interestingly, the decrease in unseen accuracy is more that that of the seen accuracy. We hypothesize that is due to the ability of the model to identify seen classes with relative ease even with just the embedding module. However discriminating between unseen classes themselves poses a bigger challenge and the statistical features further help distinguishing between seen and unseen classes.

\section{Comparison with Deep Learning Baseline: LSTM-FCN}
In this section, we study the performance of our model against another deep learning model, the Long Short Term Memory - Fully Convolutional Network \citep{karim2019insights}. We train the model using the scheme mentioned in section \ref{section: validation scheme}, utilize calibrated stacking and evaluate the metrics. Table \ref{table: LSTM-FCN} shows the results of training the LSTM-FCN.

\begin{table}[ht]
\caption{Evaluation of LSTM-FCN}
\label{table: LSTM-FCN}
\resizebox{\linewidth}{!}{%
\begin{tabular}{|c|c|c|c|c|}
\hline
Dataset Name         & AUSUC  & \(acc_s\) & \(acc_u\) & \(H\)      \\ \hline
TwoPatterns          & 52.651 & 66.973 & 37.302 & 47.916 \\
ElectricDevices      & 24.822 & 47.749 & 32.886 & 38.948 \\
Trace                & 10.666 & 6.666  & 44.00  & 11.578 \\
Synthetic Control    & 23.800  & 28.750 & 17.000 & 21.736 \\
UWaveGestureLibraryX & 32.413 & 50.418 & 39.982 & 44.492 \\
CricketX             & 13.670 & 36.752 & 27.692 & 31.585 \\
Beef                 & 11.214 & 20.145 & 12.590 & 15.495 \\
InsectWingbeatSound  & 11.577 & 32.500 & 17.000 & 22.323 \\ \hline
\end{tabular}}
\end{table}

From the table we observe that the model does not perform well for GZSL and our LETS-GZSL model is far superior. Here we hypothesize that this is due to the fact that our embedding module with contrastive loss \(E\) is better able to capture meaningful information at various time scales, than the LSTM-FCN. Similar to our ablation studies in section \ref{section: ablation studies}, the lack of time series attribute vectors which boosts LETS-GZSL's performance is also a reason why the LSTM-FCN performs worse.

\section{Symbol Table}

\begin{table}[ht]
\centering
\caption{Various symbols used and their meaning}
\label{table: symbols}
\begin{tabular}{|c|c|}
\hline
Symbol                    & Meaning                     \\ \hline
\(x_i\)                  & univariate time series         \\
\(S\)                     & number of seen classes         \\
\(U\)                     & number of unseen classes       \\
\(\mathcal{T}\)                     & total number of classes        \\
\(\mathcal{Y}_{s}\)                  & set of seen classes            \\
\(\mathcal{Y}_{u}\)                  & set of unseen classes          \\
\(\mathcal{Y_{T}}\)       & set of all classes             \\
\(E\)                     & time series embedding module   \\
\(h_i\)                  & time series embedding of \(x_i\)  \\
\(a_i\)                  & attribute vector for \(x_i\)      \\
\(q_i\)                  & concatenation of \(h_i\) and \(a_i\) \\
\(L\)                     & latent embedding module        \\
\(z_i\)                  & latent embedding of \(x_i\)       \\
\(C\)                 & classifier module              \\
\(\tau\)   & temperature parameter          \\
\(\gamma\) & calibration factor             \\ \hline
\end{tabular}
\end{table}

\end{document}